\documentclass[10pt,twocolumn,letterpaper]{article}

\usepackage{wacv}
\usepackage{times}
\usepackage{epsfig}
\usepackage{graphicx}
\usepackage{amsmath}
\usepackage{amssymb}

\usepackage{subcaption}
\usepackage{multirow}
\usepackage{color}
\usepackage{xcolor}
\usepackage{soul}
\usepackage{algorithm}
\usepackage{algorithmic}
\usepackage{caption}
\usepackage{mathtools}
\usepackage{setspace}
\usepackage{enumitem}
\usepackage{booktabs}

\definecolor{c_green}{rgb}{0.1,0.6,0.1}

\usepackage[pagebackref=true,breaklinks=true,letterpaper=true,colorlinks,bookmarks=false]{hyperref}

\wacvfinalcopy

\def\wacvPaperID{877} 

\ifwacvfinal
\usepackage[breaklinks=true,bookmarks=false]{hyperref}
\else
\usepackage[pagebackref=true,breaklinks=true,colorlinks,bookmarks=false]{hyperref}
\fi

\ifwacvfinal
\fi

\newcommand{\autoxthreed}[0]{AutoX3D }

\begin{document}

\title{Auto-X3D: Ultra-Efficient Video Understanding via Finer-Grained Neural Architecture Search}

\author{%
  Yifan Jiang$^{1}$\thanks{The first two authors contribute equally.}, Xinyu Gong$^{1*}$, Junru Wu$^{2}$, Humphrey Shi$^{3,5}$, Zhicheng Yan$^{4}$, Zhangyang Wang$^{1}$\\
  $^{1}$University of Texas at Austin~~ 
  $^{2}$Texas A\&M University ~~ \\
  $^{3}$University of Oregon~~ 
  $^{4}$Facebook AI~~ $^{5}$Picsart AI Research (PAIR)\\
  \texttt{\{yifanjiang97,xinyu.gong,atlaswang\}@utexas.edu} \\
  \vspace{-2.em}
}

\maketitle
\thispagestyle{empty}

\begin{abstract}

Efficient video architecture is the key to deploying video recognition systems on devices with limited computing resources. Unfortunately, existing video architectures are often computationally intensive and not suitable for such applications. The recent X3D work presents a new family of efficient video models by expanding a hand-crafted image architecture along multiple axes, such as space, time, width, and depth. Although operating in a conceptually large space, X3D searches one axis at a time, and merely explored a small set of 30 architectures in total, which does not sufficiently explore the space. This paper bypasses existing 2D architectures, and directly searched for 3D architectures in a fine-grained space, where \textit{block type, filter number, expansion ratio} and {attention block} are jointly searched. A probabilistic neural architecture search method is adopted to efficiently search in such a large space.   
Evaluations on Kinetics and Something-Something-V2 benchmarks confirm our \autoxthreed models outperform existing ones in accuracy up to \textbf{1.3}\% under similar FLOPs, and reduce the computational cost up to $\times$\textbf{1.74} when reaching similar performance. 

\end{abstract}

\section{Introduction}

Video models are arguably living in a much larger design space than image models, as they are expected to abstract information in both spatial- and temporal dimensions, which requires a larger set of layer types and more complicated arrangements of the layers. However, many existing video models are simply built on top of image models, which are tailored to process video by applying image models to individual/a stack of frames (e.g. two-stream network~\cite{twostreamconvnet}), replacing 2D- with 3D convolution (e.g. C3D~\cite{c3d}, I3D~\cite{i3d}, S3D~\cite{s3d}, CSN~\cite{csn}), or adding temporal convolution (e.g. R(2+1)D~\cite{r2plus1}). Such an image-to-video design process has served us as a fast path to build video architectures, but may also limit our exploration of the video architectures. On the other side, to design video architectures from scratch, one needs to take into account many more factors, than that of image recognition models. For example, one has to not only extend the inputs, features, and/or filter kernels into the temporal axis \cite{i3d}, but also strike a balance over many factors, such as the depth (number of layers), width (filter number), and spatial resolution, to achieve a good Accuracy-To-Complexity (ATC) trade-off \cite{2019efficientnet}. 

\begin{figure}[t]
    \centering
    \includegraphics[width=0.99\linewidth]{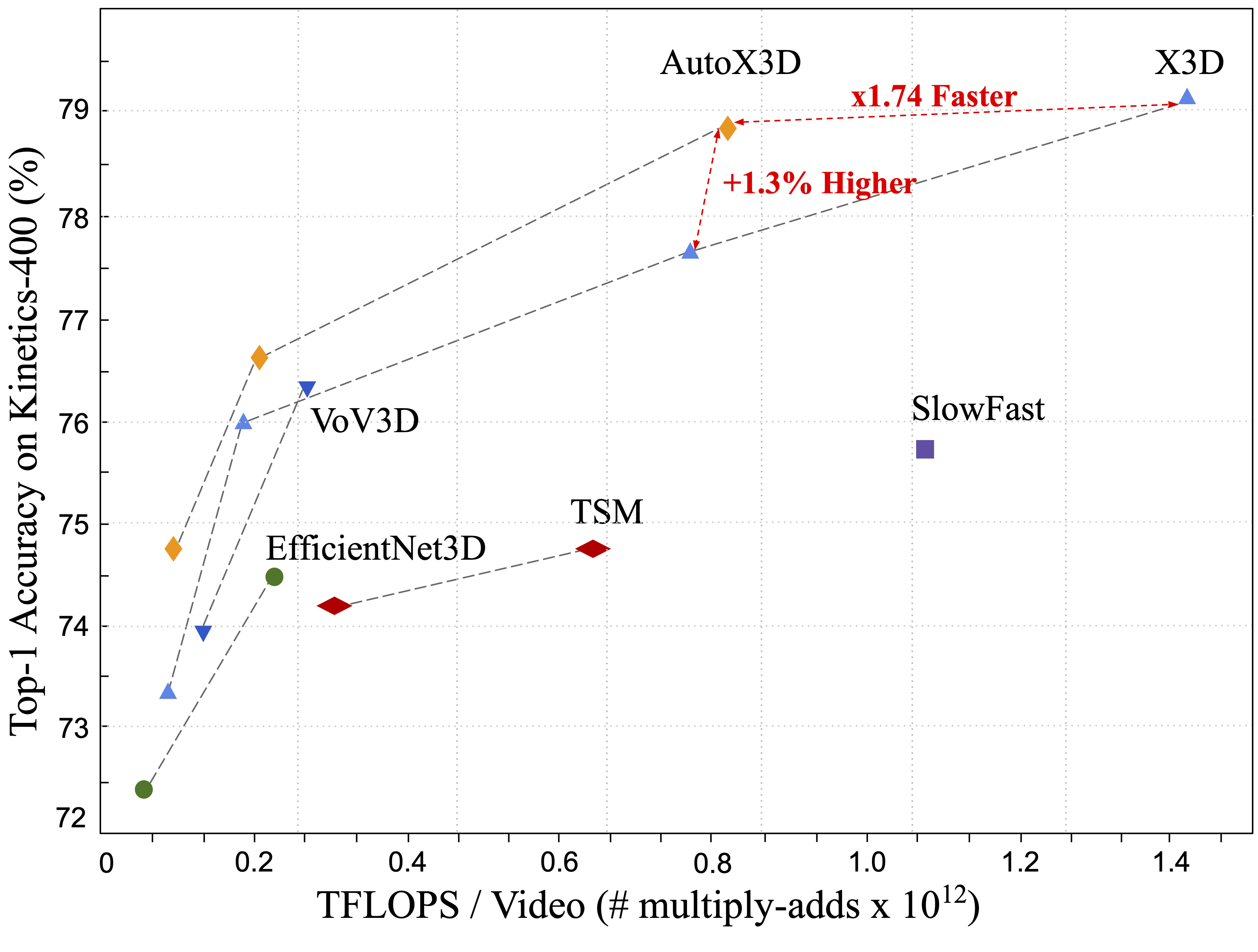}
    \caption{\textbf{Main results on Kinetics-400}. Compared to other SOTA models, the proposed AutoX3D network series achieve better Accuracy-To-Complexity (ATC) trade-off.}
    \label{fig:atc_curve}
\end{figure}

Recently, machine-designed architectures by Neural Architecture Search (NAS) have surpassed the human-designed ones for image recognition~\cite{2019AmoebaNet, pham2018efficient, 2019efficientnet, bignas}. For video action recognition, the latest work X3D~\cite{x3d} placed a new milestone in this line: it progressively expanded a hand-crafted 2D architecture into 3D spatial-temporal ones, by expanding along multiple axes, including space, time, width, and depth. As the joint search space of all axes is extremely large, X3D only searches over one axis at a time, progressively expanding the architectures to meet the target complexity. Although competitive ATC trade-off is achieved, those architectures are searched after only exploring 30 architectures in the space, which includes a total of 3,780 architectures. \textit{Can we search video architectures with better ATC if the design space is more sufficiently explored?} 

Moreover, X3D models still conform to many manual design choices in the hand-crafted 2D architecture based on MobileNet~\cite{2018mobilenetv2}. For example, the temporal-spatial kernel size is fixed to be $3\times3^2$ for depthwise convolution in the 3D MBConv (Mobile Inverted Bottleneck Conv) building block, the expansion rate is fixed to be $2.25$, and all blocks in the same stage chooses the same number of filters while the difference in the number of filters between blocks from neighboring stages is always $2\times$. For the model architecture, X3D only explores to uniformly expand the number of filters and number of blocks to change the model complexity. \textit{Can we go beyond hand-crafted 2D architectures, and directly search for 3D architectures while allowing different architectures in individual building blocks?}

This paper aims to address those questions, and takes steps further towards the direction of \textit{efficient video recognition }. The goal is fundamentally challenging as it requires us to rethink the design space of the video architecture, relax many constraints which are made to facilitate manual exploration, and more aggressively explore the design space.

To this end, this paper directly searches for efficient video architecture, leading to our work of \textbf{\autoxthreed}. Different from X3D which ignores the variations of the block types and only uniformly expands the channels, we introduce a fine-grained search space for efficient video models, by including more efficient operations. Unlike the conventional differentiable NAS (e.g. DARTS~\cite{2018darts}), which repeatedly stacks the cells of the same structure in the network, our search allows independent blocks at different layers. 
In our fine-grained space, we jointly search for the block types, filter numbers, expansion ratio, and attention blocks, substantially surpassing those in previous work on image NAS~\cite{2018darts, 2019AmoebaNet, 2019nasnet, fbnet} and video NAS~\cite{x3d, assemblenet, assemblenet, assemblenet++}. 
To support fast search, we adopt the probabilistic differentiable NAS (PARSEC)~\cite{2019pnas} as the basic search algorithm, which requires much less memory footprint than conventional DNAS (e.g. DARTS~\cite{2018darts}). Furthermore, we extend it to support the joint search among different factors (depth, channel, expansion ratio etc.). 
For better sample efficiency in our search space, we introduce the fairness-aware channel selection strategy into the probabilistic search algorithm, which enables a more stable, accurate, and scalable search framework for our task. 

To summarize, we view the contributions as following.
\begin{itemize}[leftmargin=*]
\item A fine-grained design space for efficient video recognition is proposed, to jointly search the block types, filter numbers, expansion ratio, and attention blocks.
\item A family of efficient \autoxthreed models, which are directly searched without any architecture surrogate or the use of hand-crafted image architecture, are presented. Their architectures are distinctive from hand-crafted models, and provide new insights of designing video architectures.
\item Extensive evaluations on Kinetics~\cite{kay2017kinetics} and Something-Something~\cite{sthsth} benchmarks confirm our \autoxthreed models achieve significantly better accuracy-to-complexity trade-off than other existing models.

\end{itemize}

\section{Related Work}

\subsection{Video Architecture Design}

Early video models are built on top of image ones. As the name implies, Two-Stream Network \cite{twostreamconvnet} uses two streams of image architectures to process single-frame and multi-frame optical flow inputs, respectively. To capture temporal feature, R(2+1)D~\cite{r2plus1} adds 1D temporal convolution to the 2D ResNet model while I3D~\cite{i3d} replaces 2D convolution with more expensive 3D convolution. To model long-range temporal relations, \cite{yue2015beyond,donahue2015long} runs LSTMs on top of image features extracted from video frames, while \cite{nonlocal_block} uses non-local blocks. Those video architectures are computationally heavy, and not suitable to be deployed on devices with limited computing power.

Due to the prevalence of mobile devices, designing efficient video architectures has become an increasingly important task. Efficient blocks, such as Temporal Shift module~\cite{tsm} and Video Shuffling module~\cite{ma2019learning}, are proposed to capture the temporal dynamics at low cost. More fine-grained adaptation of 2D models is also proposed. For example, in~\cite{s3d} it reports for I3D backbone, top-heavy modeling achieves better ATC trade-off by placing expensive 3D convolutions in the top layers only. However, such adaptation is not effective for ResNet backbone \cite{r2plus1}, highlighting the limitation of manually designed video architectures. Recently, search-based video architecture design, such as Tiny Video Network~\cite{piergiovanni2019tiny} and X3D~\cite{x3d}, has made significant progress towards efficient video architecture. In particular, X3D presents new insights for turning a 2D architecture into 3D one by progressively expanding it along multiple axes, such as width, depth, and time. Different from X3D, we completely search video architecture from scratch in a more fine-grained space, and demonstrate the discovered architectures can squeeze out more gain in terms of efficiency and accuracy.

\subsection{Neural Architecture Search Methods}

Neural architecture search (NAS) aims to replace the laborious human design of network architectures, as well as towards more efficient ones~\cite{2019Mnasnet}. The conventional evolution or reinforcement learning-based search methods are slow and take thousands of GPU days. For example, searching an image model for CIFAR-10 and ImageNet
required 2000 GPU days of reinforcement learning (RL)~\cite{zoph2016neural} or 3150 GPU days of evolution~\cite{2019AmoebaNet}. ENAS~\cite{pham2018efficient} introduced a parameter-sharing strategy to reduce the search time. Recent differentiable NAS (DNAS) methods~\cite{2018darts} introduced the softmax-based continuous relaxation of the architecture representation, allowing efficient search using gradient descent. However, they compute features for all layer choices, and uses much more memory than standalone model training. The probabilistic version of DNAS, such as PARSEC~\cite{2019pnas}, describes the population of architectures in the design space by a distribution, and only requires to sample one architecture at a time, which uses memory only as much as in the standalone model training. In this work, we employ PARSEC as the search procedure to search efficient video architectures in our fine-grained search space due to its low memory footprint and fast search efficiency.

\subsection{Neural Architecture Design Space}
The success of searching efficient models depends not only on the efficiency of the search method, but also on the prescription of the architecture design space. In~\cite{radosavovic2019network}, a new comparison paradigm is proposed to compare different design spaces. NASNet~\cite{2019nasnet} proposed a design space of a cell which is a directed acyclic graph. Only the connectivity between graph nodes and the operator type on each graph edge is searched. The final model is obtained by stacking cells with the same architecture. Similarly, DARTS~\cite{2018darts} also only searches the cell architecture, and repeats cells to obtain the final model. Although sharing the architectures across cells can reduce the size of model design space and simplify the search, it can be detrimental to video architecture design. Recent hand-crafted video models, such as S3D~\cite{s3d} and SlowFast~\cite{slowfast}, only use expensive 3D convolution in the top layers to improve the ATC trade-offs, while leaving convolutions in the bottom layers to be 2D. Other design choices, such as the building block type and filter number, are still manually chosen and often uniform. To search for efficient video models, we relax such hard-coded constraints, allow individual building blocks to have separate architectures and filter numbers.

\section{\autoxthreed Design Space}
\label{sec:design_space}

In this section, we introduce the design space of \autoxthreed architectures. We define a model as a stack of building blocks, and choose to use the 3D version of Mobile Inverted Bottleneck Conv block (MBConv) from MobileNetV2~\cite{2018mobilenetv2} as our building block. The original MBConv block is proven to be more efficient than the ResNet building block~\cite{resnet} in prior work~\cite{2018mobilenetv2, mobilenetv3} and its 3D version is suitable for building efficient video architectures. Unlike X3D models~\cite{x3d}, in our space, the design of the video architecture is more flexible in the following aspects.

\begin{figure*}[t!]
   \centering
   \includegraphics[width=0.95\textwidth]{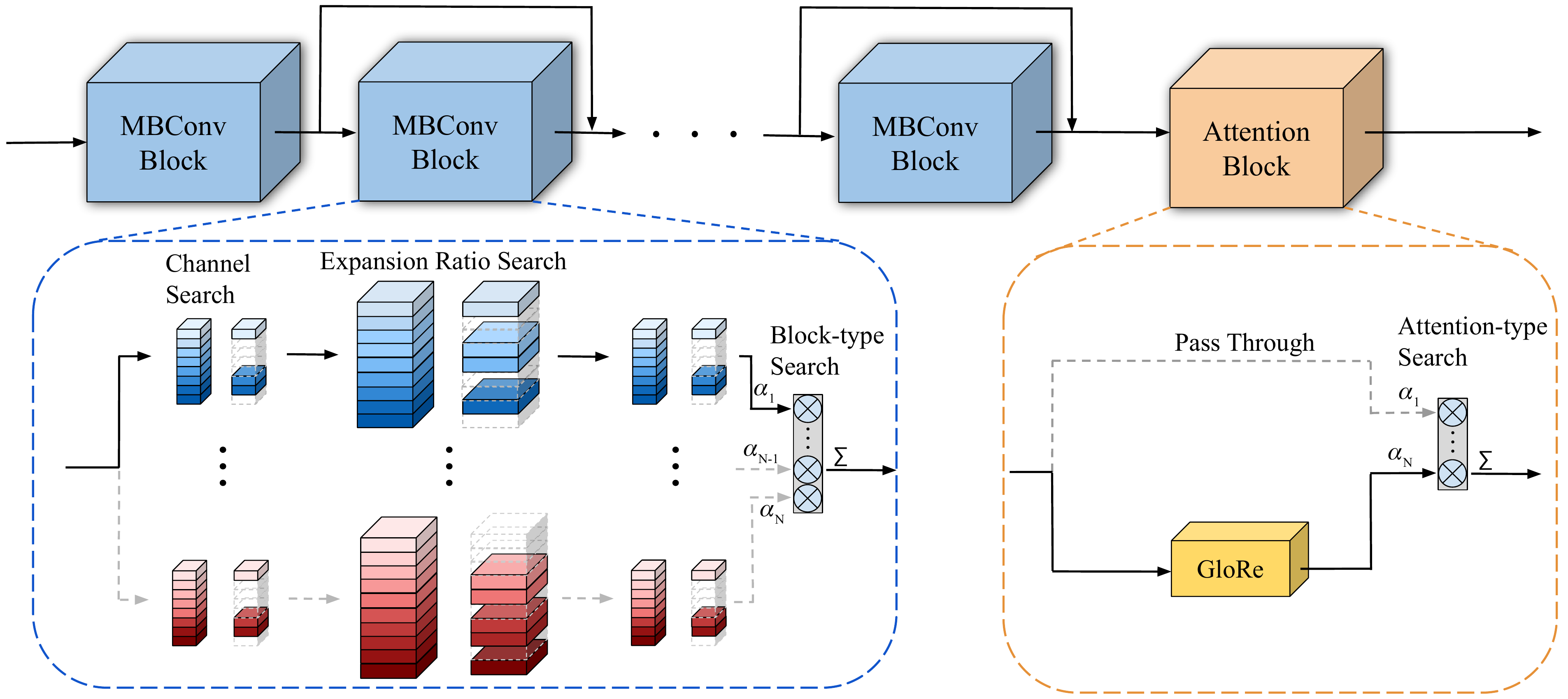} 
   \caption{\textbf{The macro-architecture of \autoxthreed design space.}}
   \label{fig:design_space}
\end{figure*}

\subsection{Nonuniform Block Micro-Architecture}
In X3D models, the 3D MBConv building block always consists of a conv$1\times1$ layer to expand the filter number, a depthwise convolution layer of kernel size $3\times3^2$ to convolve with spatial features, and another conv$1\times1$ layer to shrink the filter number in the final output feature. Depthwise convolution with kernel $3\times3^2$ is more expensive than that with a 2D kernel of size $1\times3^2$, and such choice of uniform building block micro-architecture is sub-optimal to achieve good ATC. Many prior works~\cite{r2plus1, s3d} have examined the progressively varying temporal-spatial feature patterns along with the model depth. For example, S3D~\cite{s3d} work has shown employing 3D convolution at bottom layers (close to input) is less cost-effective than using it at top layers (close to final prediction). On the other side, other convolution kernel sizes, such as $1\times5^2$ and $3\times5^2$, are not yet explored in hand-crafted models due to a large number of different combinations of spatial-temporal kernel size. In \autoxthreed design space, we consider 3D MBConv block with a different spatial-temporal kernel size in depthwise convolution as a separate block micro-architecture.


In Table~\ref{tab:mbconv3d_micro_arch}, we present the choices of the 3D MBConv block micro-architectures in our design space.

\begin{table}[t]
\centering
\resizebox{0.3\linewidth}{!}{

  \begin{tabular}{c|cc}
  \toprule
    Type & Kernel  \\
    \specialrule{.1em}{.1em}{.1em}   

    t1\_s3 & $1\times3^2$  \\   
    t1\_s5 & $1\times5^2$ \\  
    t3\_s3 & $3\times3^2$ \\   
    t3\_s5 & $3\times5^2$ \\
    t5\_s3 & $5\times3^2$  \\   
    t5\_s5 & $5\times5^2$ \\
    \bottomrule

  \end{tabular}
}
\caption{\textbf{The choices of MBConv3D block micro architecture in our space.}}
\label{tab:mbconv3d_micro_arch}
\end{table}

\subsection{Searchable Filter Number}

The choices of the filter number in the building blocks of the hand-crafted models often follow simple heuristics. For example, in I3D~\cite{i3d}, S3D~\cite{s3d}, R(2+1)D~\cite{r2plus1}, and X3D~\cite{x3d} models, all of which have 4 stages and each stage has multiple building blocks, the filter number of the output feature from the building blocks is fixed within the stage, and is doubled in the next stage. The principle of this heuristic choice is originally from 2D ResNet work~\cite{resnet}, whereby doubling the filter number when the spatial resolution of the 2D feature map is reduced by half in the next stage, all building blocks use a similar amount of FLOPS and the compute of the overall model is evenly distributed to the building blocks. 

However, this principle does not hold for 3D video models, as the 3D feature map has both temporal- and spatial-resolution, and often only the spatial-resolution is reduced by half in the next stage while the temporal resolution is not reduced for achieving high recognition performance. For example, in SlowFast~\cite{slowfast} and X3D~\cite{x3d} models, the temporal resolution is not reduced in all 4 stages.

Moreover, the choice of evenly distributing computation to all building blocks is quite ad-hoc. In our design space, we do not manually prescribe the filter number for the building blocks. Instead, we design a separate range of filter number choices for individual building blocks in the model, and even allow building blocks in the same stage to have different filter numbers. As the choices of filter number have a large impact on both the recognition performance and model computational cost, our flexible design and fine-grained choices allow the search methods to discover architectures with a better ATC trade-off.

The choices of filter numbers for individual building blocks are shown as part of the macro-architecture of our design space in Table~\ref{tab:macro-arch}.

\subsection{Nonuniform Expansion Rate}
In MBConv block, a conv$1\times1$ layer is used to expand the filter number, and the expansion rate decides the channel number of the feature map where the following depthwise convolution will operate on. For simplicity, a uniform choice of the expansion rate is often chosen by the hand-crafted models. For example, in the original MobileNet family of models~\cite{2018mobilenetv2, mobilenetv3}, all MBConv blocks use the expansion rate 6 while in X3D models~\cite{x3d}, the expansion rate is always 2.25. The uniform choice of the expansion rate ignores the variations of feature representation needed at different model depths, and we hypothesize it leads to sub-optimal architectures. 

In our \autoxthreed design space, we predefine a wide range of choices for the expansion ratio in individual building blocks, and allow different choices are chosen by the building blocks. The choices of the expansion ratio can be seen in the macro-architecture of our design space in Table~\ref{tab:macro-arch}.

\begin{table}[!t]
\centering
\centering
{\small
\resizebox{\linewidth}{!}{
\setlength{\tabcolsep}{1.2mm}{
    \begin{tabular}{lcccccc}
    \toprule
    Max Input & Block & \multirow{2}{*}{Expansion} & \multirow{2}{*}{Channel} & \multirow{2}{*}{Number} & Spatial & Att.   \\
    $C\times T \times S^2$ & Type &  &  &  &  Stride & Block \\
  \specialrule{.15em}{.1em}{.1em}      
    3$\times$$13$$\times$ $160^{2}$ & data & 1 & 24 & 1 & 1 & -\\ \midrule
    3$\times$$13$$\times80^{2}$ & 1$\times3^{2}$ & 1 & 24 & 1 & 2 & -\\ \midrule
    28$\times$$13$$\times$ $40^{2}$ &  TBS  &  \multirow{10}{*}{ 1.5 $\sim$  6.0, 0.75 }   & 12 $\sim$  28, 4 & 1 & 2  & TBS \\
    28$\times$$13$$\times$ $40^{2}$ & TBS &  & 12 $\sim$ 28, 4 & 2 & 1 & TBS \\
    64$\times$$13$$\times$ $20^{2}$ & TBS &  & 24 $\sim$ 64, 8 & 2 & 2 & TBS \\
    64$\times$$13$$\times$ $20^{2}$ & TBS &  & 24 $\sim$ 64, 8 & 3 & 1 & TBS \\
    132$\times$$13$$\times$ $10^{2}$ & TBS &  & 48 $\sim$ 132, 12 & 2 & 2 & TBS \\
    132$\times$$13$$\times$ $10^{2}$ & TBS &  & 48 $\sim$ 132, 12 & 3 & 1 & TBS \\
    132$\times$$13$$\times$ $10^{2}$ & TBS &  & 48 $\sim$ 132, 12 & 3 & 1 & TBS \\
    132$\times$$13$$\times$ $10^{2}$ & TBS &  & 48 $\sim$ 132, 12 & 3 & 1 & TBS \\
    264$\times$$13$$\times$ $5^{2}$ & TBS &  & 96 $\sim$ 264, 24 & 2 & 2 & TBS \\
    264$\times$$13$$\times$ $5^{2}$ & TBS &  & 96 $\sim$ 264, 24 & 2 & 1 & TBS \\
    264$\times$$13$$\times$ $5^{2}$ & TBS &  & 96 $\sim$ 264, 24 & 3 & 1 & TBS \\
    \midrule
    432 & pool & - & 432 & 1 & - & - \\ \hline
    2048 & fc & - & 2048 & 1 & - & -\\
    \bottomrule
    \end{tabular}
    }
  }
}
\caption{\textbf{The macro-architecture of our \autoxthreed design space. ``TBS'' means the operators that are to be searched.}  }
\label{tab:macro-arch}
\end{table}

\subsection{Searchable Attention Block}
\label{sec:attention_block_variable}

Attention blocks, such as Nonlocal-~\cite{nonlocal_block} and GloRe block~\cite{glore}, can be easily plugged into the backbone model for improving the performance. One can choose to insert more than one such blocks at different layers of the backbone. For example, in the GloRe work~\cite{glore}, for ResNet-50 backbone, it chooses to insert 3 GloRe blocks at Res4 stage for achieving a large gain in accuracy. Since modern deep models often contain dozens of layers and we can insert the attention block after any layer, a manual exhaustive exploration for selecting the best subset of layers to insert attention blocks is almost intractable. In our design space, we view the insertion of the attention block as a search problem, and define a set of choices of the attention block below to search the inserted position.

\noindent{\textbf{Pass through.}} The feature will simply pass through the block without any computation. In practice, it represents no attention block placed here.

\noindent{\textbf{GloRe.}} The original GloRe block, which includes a feature projection module, a graph convolution network (GCN) module, and an inverse feature projection module. The feature projection module projects spatial features from the coordinate space to the interaction space by aggregating spatial features with attention. The GCN module reasons over the projected features nodes in the interaction space. Finally, the inverse projection module re-projects the feature nodes back into the coordinate space.



\subsection{Final \autoxthreed Design Space}

The final design space is illustrated in Figure~\ref{fig:design_space}. In Table~\ref{tab:macro-arch}, we prescribe a video macro-architecture, which mainly consists of a stack of 3D MBConv building blocks. We organize the blocks by groups, and blocks in each group share the same design choices. For each group, we search the micro-architecture, filter number, the expansion rate, and the type of attention block which will be inserted after each group. We allow the block groups to have different choices along those design axes.  In total, our design space contains $2\times10^{32}$ different video architectures, and represents a fine-grained space at a scale that has not been explored before.

\section{Search Method}

\subsection{Probabilistic-based Architecture Search}
As video models commonly require larger computation resources than image, directly applying the original differentiable NAS becomes unpractical. Here we adopt the PARSEC~\cite{2019pnas} approach, a probabilistic version of the differentiable NAS method, to guide our search procedure. Compared to DARTS \cite{2018darts}, it only requires much memory as is needed to train a single architecture from our search space. This is because of a memory-efficient sampling procedure as we
learn a probability distribution over high-performing neural network architectures. Here we define a unique discrete architecture as $A$, which is sampled from a prior distribution $ P(A|\pmb{\alpha})$. Architecture parameters $\pmb{\alpha}$ denote the probabilities of choosing different operations. The goal of PARSEC is to optimize the architecture parameter $\pmb{\alpha}$, in order to maximize the architecture accuracy. Concretely, for video recognition where we have video samples $\mathbf{X}$ and labels $\mathbf{y}$, probabilistic NAS can be formulated as optimizing the continuous architecture parameters $\alpha$ via an empirical Bayes Monte Carlo procedure~\cite{rasmussen2003bayesian}
\begin{equation}
\begin{split}
P(\mathbf{y} | \mathbf{X},   \mathbf{\omega}, \alpha ) & = \int P(\mathbf{y} | \mathbf{X}, \mathbf{\omega}, A) P(A|\pmb{\alpha}) \textup{d}A \\
& \approx \frac{1}{K} \sum_{k} P(\mathbf{y} | \mathbf{X}, \mathbf{\omega}, A_{k}),
\end{split}
\end{equation}
where $\omega$ denotes the model weights. The continuous integral of data likelihood is approximated by sampling $K$ architectures and averaging the data likelihoods from them. We can jointly optimize architecture parameters $\pmb{\alpha}$ and model weights $\mathbf{\omega}$ by estimating gradients $\nabla _{\alpha}\;\textup{log}\;P(\mathbf{y}|\mathbf{X}, \omega, \pmb{\alpha} )$ and $\nabla _{w} \;\textup{log}\;P(\mathbf{y}|\mathbf{X}, \omega, \pmb{\alpha})$ through the sampled architectures. Typically, the number of sampled architecture $K$ is set to 13, which is sufficient to search for a good architecture empirically, according to our preliminary experiments.

To target at efficient architecture discovering, we rebuild PARSEC to support cost-aware search that aiming to not only finding the best performance architecture but also under determined FLOPs constraint. We achieve this goal by adding another cost-aware hinge loss term to the original loss function. Given an architecture $A_{k}$, the cost-aware loss term $c$ can be written as:
\begin{equation}
\label{eq:cost}
    c(A_{k}) = \frac{1}{T} \max(FLOPs(A_{k})-T, 0),
\end{equation}
where $T$ is our target FLOPs.
All the search variables are jointly searched. 

\subsection{Fair Channel Selection}
\label{sec:fair_channel_selection}
\begin{figure*}[t!]
\begin{center}
 \includegraphics[width=\linewidth]{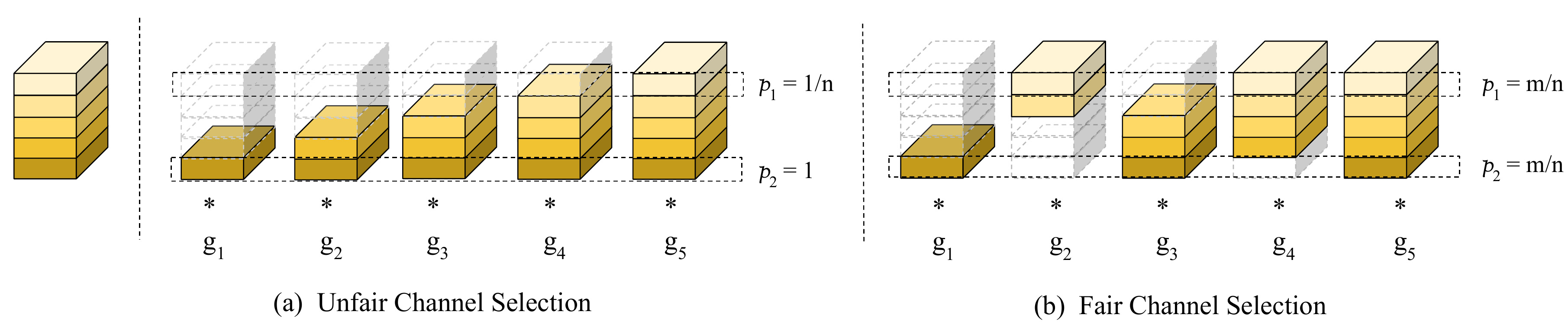}
\end{center}
  \caption{\textbf{Illustration of fairness-aware channel search}, using five channel parts for example. Where $p_i$ is the initial probability of each filter and $g_{i}$ represent the architecture weight of each candidates. We only activate one candidates in the forward pass. All channel candidates share weight from the original filters. For unfair channel selection, the small probability one gets fewer chance to update the corresponding filter compared to large probability one.}
\label{fig:fair}
\end{figure*}

An important pitfall we found during channel search is the \textit{unfairness} for the chances of candidate channels being selected. We show an example in Figure~\ref{fig:fair}: on the left, when splitting the super kernel into $N$ parts ($N$ denotes the channel space size, with $N$ = 5 in the figure just as an example) in the default order, the bottom part will be shared with all channel candidates and therefore always be updated. Meanwhile, the higher bottoms have fewer chances to be updated, with the top one only receiving sparse updates of 1/$N$ probability. That unfairness often leads to the supernet training collapsing onto the most frequently updated channel(s) (e.g. the bottom one), and mislead the architecture selection to sub-optimal performance. To tackle this issue, we propose to \textit{select channels with fairness}, i.e., the assignment of channel candidates shall ensure the chance of selecting/updating each super kernel part to be as equal as possible. Figure~\ref{fig:fair} right displays the fairness-aware selection pattern for $N = 5$, where each part now has the equal chance of 3/5 to be selected and updated. A more general discussion on how to design such fair patterns for different $N$s can be found in the supplement.

\section{Experiments}

\subsection{Implementation Details}
\subsubsection{Datasets}

We adopt two benchmarks of video action recognition in the experiments, including Kinetics~\cite{kay2017kinetics} and Something-Something-V2~\cite{sthsth}.

\noindent{\textbf{Kinetics~\cite{kay2017kinetics}}}. Kinetics-400 is a large-scale dataset that contains 400 action classes with 240K training examples and 20K validation examples. To accelerate searching, we adopt Mini-Kinetic200~\cite{xie2018rethinking} as our searching dataset, which is a subset of Kinetics-400 containing 200 classes of videos. As for the evaluation of the searched architecture, we use the full standard Kinetics-400 dataset for training and testing.


\noindent{\textbf{Something-Something-V2}}.  Something-something-V2 dataset contains 108k videos with 174 classes of diverse actions, while each video lasting between 2 to 6 seconds. Different from Kinetics\cite{kay2017kinetics}, the video sequence in something-something is more temporal related, as it focuses on humans performing predefined basic actions with different objects. Therefore, this dataset serves as a suitable benchmark to evaluate the effectiveness of temporal modeling. 


\subsubsection{Search and Derivation Setting}
\noindent{\textbf{Architecture Search}} We implement PARSEC~\cite{2019pnas} in PyTorch, and use 64 Nvidia V100 GPUs to search architectures on Mini-Kinetics-200. By default, we use the input video clip of size $T\times S^2 = 13\times160^2$ with a spatial scale jittering range of [182, 228], and set the target FLOPS to $2G$ FLOPS, which is on par with the FLOPS of X3D-S model. We adopt Adam with 0.02 learning rate and 0 weight decay to optimize the architecture weights. SGD is adopted to update the supernet's parameters with a learning rate of 0.6, 0.9 momentum, and 5e-5 weight decay. The total search process takes 600 epoches with 3 days. 

\noindent{\textbf{Architecture Training}} We evaluate the searched architectures by using the open-source ClassyVision~\cite{classyvision} framework to train them from scratch on the benchmarks. We use 64 Nvidia V100 GPUs in the experiments. We adopt SGD with 0.9 momentum and 5e-5 weight decay. The learning rate is set to 0.4 following the half-period cosine decaying strategy and mini-batch size is 8 clips per GPU. The whole training process takes 300 epochs while we use linear warm-up strategy \cite{goyal2017accurate} for the first 34 epochs. 
Dropout is adopted at the head of the network with the probability of 0.5.  We also apply AutoAugment~\cite{autoaugment} on each frame of input video clip. The data pre-processing part is the same as the searching part. More training detials are illustrated in the supplement.

\subsubsection{Architecture Evaluation}
To be comparable with previous work, we mainly follow the evaluation setting of X3D~\cite{x3d}. We evaluate the searched architecture on the validation set. We by default adopt 30-views evaluation, unless specifically stated. Concretely, in 30-views evaluation, 10 clips are uniformly sampled temporally from each video first. Each clip is cropped out using LeftCenterRight cropping strategy~\cite{x3d}, which covers the longer axis of the original clip. The final prediction of a video is obtained by averaging the predictions for all corps of each clip. More details can be found in our supplement.

\subsection{Ablation Study}
\subsubsection{Searching for Attention Block}
\begin{table}[t!]
\centering
\resizebox{\linewidth}{!}{

  \begin{tabular}{c|cccc|c|c|c}
  \toprule
     \multirow{2}{*}{Model} &  \multicolumn{4}{c|}{Attention location}  & Params & FLOPS & \multicolumn{1}{c}{Accuracy ($\%$)} \\
    & S1 & S2 & S3 & S4 & (M) & (G) & Top-1 \\
    \specialrule{.1em}{.1em}{.1em}
    X3D-S  & - & - & - & - & 3.4 & 1.96 & 72.9  \\ \hline
     \multirow{4}{*}{Manual}  & 2 & - & - & -  & 3.5 & 2.33 & 72.6 \textcolor{purple}{(\textbf{$-$0.3})} \\
     & - & 2 & - & - & 3.6 & 2.56  & 73.0 \textcolor{c_green}{(\textbf{$+$0.1})} \\
     & - & - & 2 & - & 3.7 &  2.39 & 72.8 \textcolor{purple}{(\textbf{$-$0.1})} \\
     & - & - & 2 &  3 & 5.4 & 2.92  &  73.0 \textcolor{c_green}{(\textbf{$+$0.1})} \\ \hline
    Automatic & \multicolumn{4}{c|}{Searched} & 4.1 & 2.44  &  \textbf{73.3} \textcolor{c_green}{(\textbf{$+$0.4})}  \\
\bottomrule

  \end{tabular}
}
\vspace{-3pt}
\caption{\textbf{Ablation in attention block insertion location}. The number $k$ under column S$n$ denotes $k$ attention blocks are uniformly placed in stage $n$. 
}
\label{tab:search_attention}
\vspace{-5pt}
\end{table}

To ablate the importance of attention block search, we conduct an ablation study based on the original X3D-S~\cite{x3d} backbone by 1) Manually insert attention block in each stage; 2) Automatically search the suitable position of attention block through probabilistic search algorithm~\cite{2019pnas}. We then compare them with the original baseline (X3D-S) on Kinetics-400 testbed. 

As shown in Table \ref{tab:search_attention}, the first row represents the X3D-S baseline, reaching 72.9\% top-1 accuracy with 1.96 GFLOPs cost. The second row includes four different architectures, where the attention blocks are inserted to different positions, ranging from the shallow layer (stage 1) to the deep layer (stage 4). These limited manual attempts bring minor improvement ($+$\textbf{0.1\%}) based on the X3D-S backbone and sometimes even hurt the performance ($-$\textbf{0.3\%}), motivating us to apply the automatic tools (\eg, NAS) to reap the performance benefits of various position choices. We finally adopted the probabilistic-based architecture search on discovering the suitable position of attention blocks, where a block-wise position search is conducted with the rest of the backbone being fixed. To this end, the searched architecture reaches 73.3\% top-1 accuracy, surpassing the original X3D-S baseline by a notable margin (\textbf{$+$0.4\%}) with a comparable FLOPs cost.

\subsubsection{Evaluation of Fairness-Aware Search}
\label{sec:exp_fairness_filter_number}
The default channel search simply split the super kernel into $N$ splits and sample the candidate by default order. As we discussed in Sec.~\ref{sec:fair_channel_selection}, this naive selecting strategy will make some parts of the super kernel to be over-sampled and updated unfairly. Thus the discovered model will easily collapse to the candidate which is updated more frequently. 

\begin{figure}[t!]
\vspace{-3pt}
    \centering
\includegraphics[width=\linewidth]{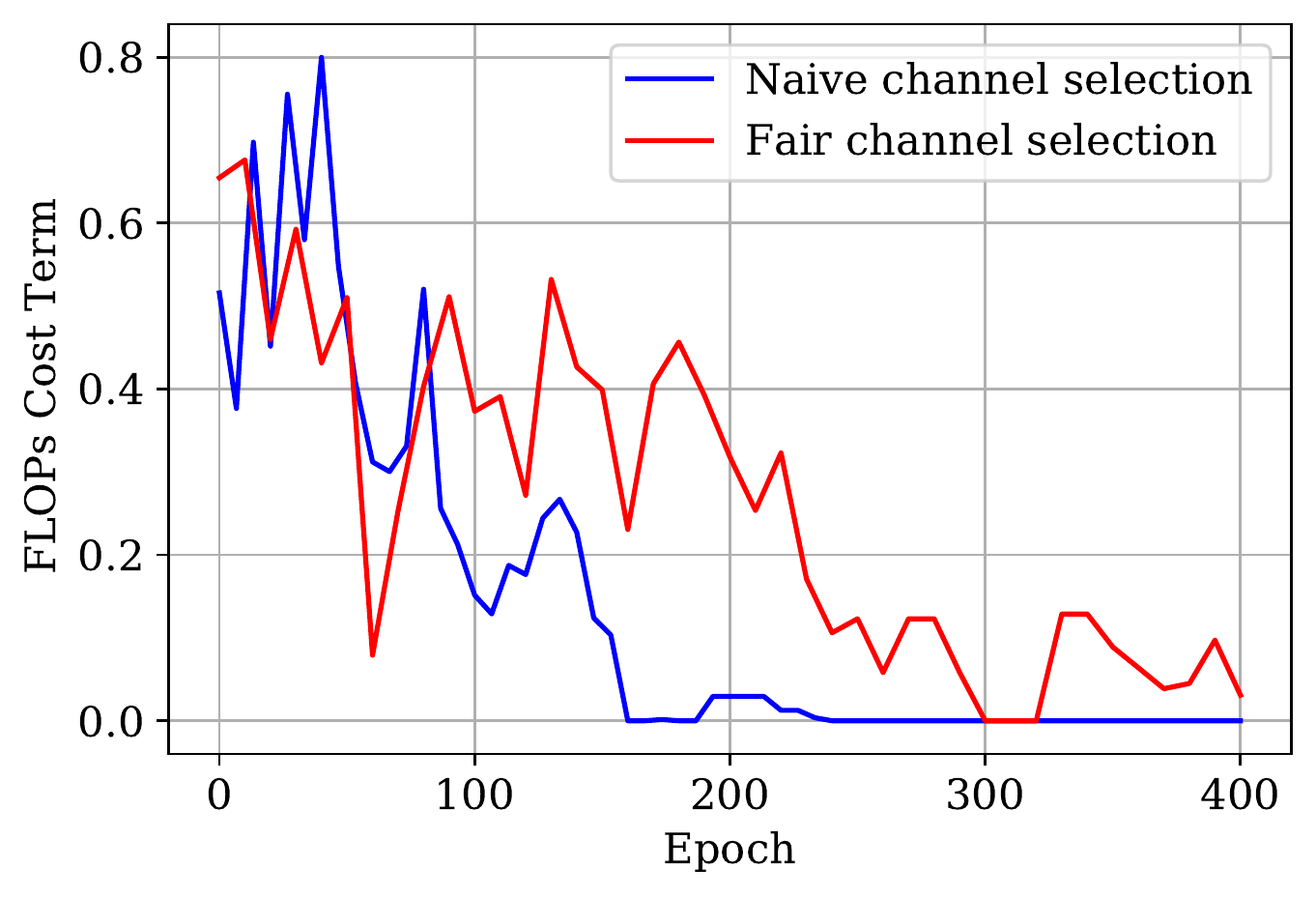}   
\vspace{-8pt}
\caption{\textbf{FLOPs during searching.} For each epoch (x-axis), we derive the most probable architecture and measure its FLOPs (y-axis). The target FLOPs $T$ (Equation \ref{eq:cost}) is set to 2.0 GFLOPs.}
    \label{fig:flops_curve}
\end{figure}

To better understand the effectiveness of the proposed fairness-aware sample strategy, we conduct experiments that only search for the output channel and expansion ratio by separately using these two sampling strategies, while other factors are fixed.
As shown in Figure \ref{fig:flops_curve}, the blue curve represents the FLOPs cost of discovered architectures during the search process by using the naive channel selection strategy, and the red curve represents the fairness-aware one. Given a target FLOPs, the naive channel selection strategy easily collapses to the smaller subnets, due to the smaller channel candidates are updated more frequently than others in each super kernel. In contrast, the fairness-aware strategy can discover a more suitable subnet that is closer to the target FLOPs cost. The collapsed discovered architecture, shows poor performance after evaluation, with a large margin ($-$\textbf{2.6\%}) behind the well-discovered fairness-based one. Details are shown in Table \ref{tab:cost}.

\begin{table}[t!]
\centering
\resizebox{\linewidth}{!}{
  \begin{tabular}{l|c|c|c}
  \toprule
  \multirow{2}{*}{Search Strategy} & Target & Discovered  & Accuracy (\%) \\ 
   & FLOPs (G) & FLOPs (G) & Top-1 \\\specialrule{.1em}{.1em}{.1em}
  Naive Channel Selection & \multirow{2}{*}{2.0} & 1.7\textcolor{purple}{($-$\textbf{0.3})} & 70.5 \textcolor{purple}{($-$\textbf{2.6})} \\
  Fair Channel Selection (\textbf{ours}) &  & \textbf{2.0} & \textbf{73.1}  \\
  \bottomrule
  \end{tabular}
}
  \caption{\textbf{Architecture search with different channel selection mechanisms.}}
\label{tab:cost}
\end{table}

\subsection{Comparing \autoxthreed Models with Others}
\label{sec:comparison_sota}
In this section, we compare our searched \autoxthreed models with other SOTA models. 
\vspace{-1.em}
\subsubsection{Results on Kinetics}
\begin{table}[!t]
\centering
{\small
\resizebox{\linewidth}{!}{
\setlength{\tabcolsep}{1mm}{
    \begin{tabular}{l|c|c|c|c|c}
    \toprule
    Model & Pretrain & Top-1 & Frames & GFLOPs$\times$Views & Param.\\
  \specialrule{.15em}{.1em}{.1em}      
    I3D \cite{carreira2017quo} & ImgNet & 71.1 & 64 & 108 $\times$ N/A & 12.0 \\
    2-Stream I3D \cite{carreira2017quo} & ImgNet  & 75.7 & 64 & 216 $\times$ N/A & 25.0 \\
    MF-Net \cite{chen2018multi} & ImgNet  & 72.8 & 16 & 11.1 $\times$ 50 & 8.0 \\
    TSM R50 \cite{tsm} & ImgNet  & 74.7 & 16 & 65.0 $\times$ 10 & 24.3 \\
    S3D-G \cite{xie2018rethinking} & - & 74.7& 64 & 71.0 $\times$ N/A & N/A \\
    
     
    2-Stream I3D \cite{carreira2017quo} & - & 71.6 & 64 & 216 $\times$ N/A & 25.0 \\
    R(2+1)D \cite{tran2018closer} & - & 72.0 & 32 & 152 $\times$ 115 & 63.6 \\
    2-Stream R(2+1)D \cite{tran2018closer} & - & 73.9 & 32 & 304 $\times$ 115 & 127 \\
    ip-CSN-50 \cite{csn} & - & 70.8 & 8 & 11.9 $\times$ 30 & 14.3 \\
    ip-CSN-101 \cite{csn} & - & 71.8 & 8 & 15.9 $\times$ 30 & 24.5 \\
    \midrule
    TSM Mb-V2 \cite{tsm} & ImgNet & 69.5& 16 & 6.0 $\times$ N/A & 3.9 \\
    VoV3D-M~\cite{lee2020diverse} & - & 73.9 & 16 & 4.4 $\times$ 30 & 3.8 \\
    X3D-S~\cite{x3d} & - & 73.3 & 13 & 2.7 $\times$ 30 & 3.8 \\
    
    \textbf{AutoX3D-S} & - & \textbf{74.7} & 13 & 2.9 $\times$ 30 & 3.5 \\

    \midrule
    MobileNetV2-3D (impl.)  & - & 71.2 & 16 & 6.5 $\times$ 30 & 4.4 \\
    SlowFast 4×16, R50 \cite{slowfast} & - & 75.6 & 16 & 36.1 $\times$ 30 & 34.4 \\
    VoV3D-L~\cite{lee2020diverse} & - & 76.3 & 16 & 9.3 $\times$ 30 & 6.2 \\
    X3D-M~\cite{x3d} & - & 76.0 & 16 & 6.2 $\times$ 30 & 3.8 \\
    \textbf{AutoX3D-M} & - & \textbf{76.7}& 16 & 6.8 $\times$ 30 & 3.5 \\
    \midrule
    SlowFast 8×8, R101 \cite{slowfast} & - & 77.9 & 8 & 106 $\times$ 30 & 53.7 \\ 
    X3D-L & - & 77.5 & 16 & 24.8 $\times$ 30 & 6.1 \\
    X3D-XL & - & 79.1 & 16 & 48.4 $\times$ 30 & 11.1 \\
    \textbf{AutoX3D-L} & - & \textbf{78.8} & 16 & \textbf{27.8} $\times$ 30 & 6.2 \\
    \bottomrule
    \end{tabular}
    }
  }
}
\vspace{-0.5em}
\caption{\textbf{Comparison with the state-of-the-art methods on Kinetics-400 dataset}. We use (GFLOPs $\times$ views) to represents inference cost $\times$ number of views).}
\label{tab:kinetics_sota}
\vspace{-10pt}
\end{table}

We mainly follow the comparison convention of X3D \cite{x3d} on popular large-scale benchmarks, and compare the performance of our AutoX3D with current state-of-the-art efficient architectures. 
The Top-1 accuracy as well as the FLOPS cost are reported on Kinetics-400 dataset. We divide the models into four categories according to their FLOPs cost, shown in Table \ref{tab:kinetics_sota}. The first group includes previous high-performance yet inefficient methods. Those methods either cost more FLOPs and parameters compared to ours, or get worse performance. In the second group, the proposed AutoX3D-S architecture achieves 74.7\% accuracy, better than TSM with mobilenet backbone ($+$\textbf{5.2\%}), VoV3D-M ($+$\textbf{0.8}\%), and X3D-S ($+$\textbf{1.4}\%). Regarding FLOPs, AutoX3D-S costs $\times$\textbf{2} smaller FLOPs than TSM with mobilnet backbone, $\times$\textbf{1.5} smaller FLOPs than VoV3D-M, and comparable FLOPs compared to X3D-S. In the third group,
we evaluate a larger architecture, named AutoX3D-M, which is derived from AutoX3D-S by increase the input frame number to 16 and input video spatial resolution to 256. 
It achieves 76.7\% accuracy, largely outperforms our implemented MobileNetV2-3D ($+$\textbf{5.5\%}) and SlowFast 4x16 ($+$\textbf{1.1\%}), respectively. Compared to X3D-M, AutoX3D-M obtains $+$\textbf{0.7\%} higher accuracy with similar FLOPS cost. In the fourth group, we present our biggest architecture AutoX3D-L by scaling the network depth of AutoX3D-M to $\times$2 deeper and increasing the input video spatial resolution to 356. AutoX3D-L surpasses X3D-L by $\textbf{1.3\%}$ accuracy with similar FLOPs cost, and reaches comparable performance compared to X3D-XL with much smaller ($\times$\textbf{1.74}) FLOPs cost.
\vspace{-1.em}
\subsubsection{Results on Something-Something-V2}
\begin{table}[t!]
\centering
{\small
\resizebox{\linewidth}{!}{
\setlength{\tabcolsep}{1mm}{
    \begin{tabular}{l|c|c|c|c|c}
    \toprule
    Model & Pretrain &  Top-1  & Frames & GFLOPs & Param. (M)\\
  \specialrule{.15em}{.1em}{.1em}   
    TSM$_{8F}$ \cite{tsm} & \multirow{11}{*}{ \rotatebox[origin=c]{90}{ImageNet}   } & 59.1 & 8 & 32.9 & 23.9 \\
    TSM$_{16F}$ \cite{tsm} &  & 63.4 & 16 & 65.8 & 23.9 \\
    STM$_{8F}$ \cite{stm} &  & 62.3  & 8 & 33.3 & 24.0 \\
    STM$_{16F}$ \cite{stm} &  & 64.2 & 16 & 66.5 & 24.0 \\
    GST${_8F}$ \cite{GST}  &  & 61.6  & 8 & 35.4 & - \\
    GST${_16F}$ \cite{GST} &   & 62.6  & 16 & 35.4 & - \\
    I3D + STIN + OIE \cite{STIN} &  & 60.2  & - & - & - \\
    Dynamic Inference \cite{wu2020dynamic} &  & 58.2  & - & 35.4 & - \\
    SmallBigNet$_{8F}$ \cite{SmallBigNet} &  & 61.6 & 8 & 52 & - \\
    SmallBigNet$_{16F}$ \cite{SmallBigNet} &  & 63.8  & 16 & 105 & - \\
    \hline
    X3D-S & \multirow{4}{*}{-} & 61.3 & 13 & 2.0 & 3.8 \\
    X3D-M &  & 62.7 & 16 & 4.7 & 3.8 \\
    \textbf{AutoX3D-S} &  & 62.1 & 13 & \textbf{2.2} & 3.5 \\
    \textbf{AutoX3D-M} &  & \textbf{63.4} & 16 & \textbf{5.3} & 3.5 \\
    \bottomrule
    \end{tabular}
    }
  }
}
\caption{\textbf{Comparison with the state-of-the-art methods on Something-Something-V2 dataset}. }
\label{tab:sthsthv2_sota}
\end{table}
In addition, we directly transfer the discovered architectures on Something-something v2 dataset. We include current state-of-the-art methods in Table \ref{tab:sthsthv2_sota}, where AutoX3D model family achieves competitive performance \textbf{without} the necessity of pre-training. For example,
AutoX3D-S achieves similar accuracy (62.1\% vs 62.3\%) with STM$_{8F}$ while the FLOPs cost is $\times$\textbf{15} smaller. The Top-1 accuracy of AutoX3D-M is on par with SmallBigNet$_{16F}$ (63.4\% vs 63.8\%), while the FLOPs cost is \textbf{$\times$20} fewer.

\subsection{Understanding \autoxthreed Architectures}
\label{sec:understand_autox3d}
\begin{figure}[t]
    \centering
\includegraphics[width=\linewidth]{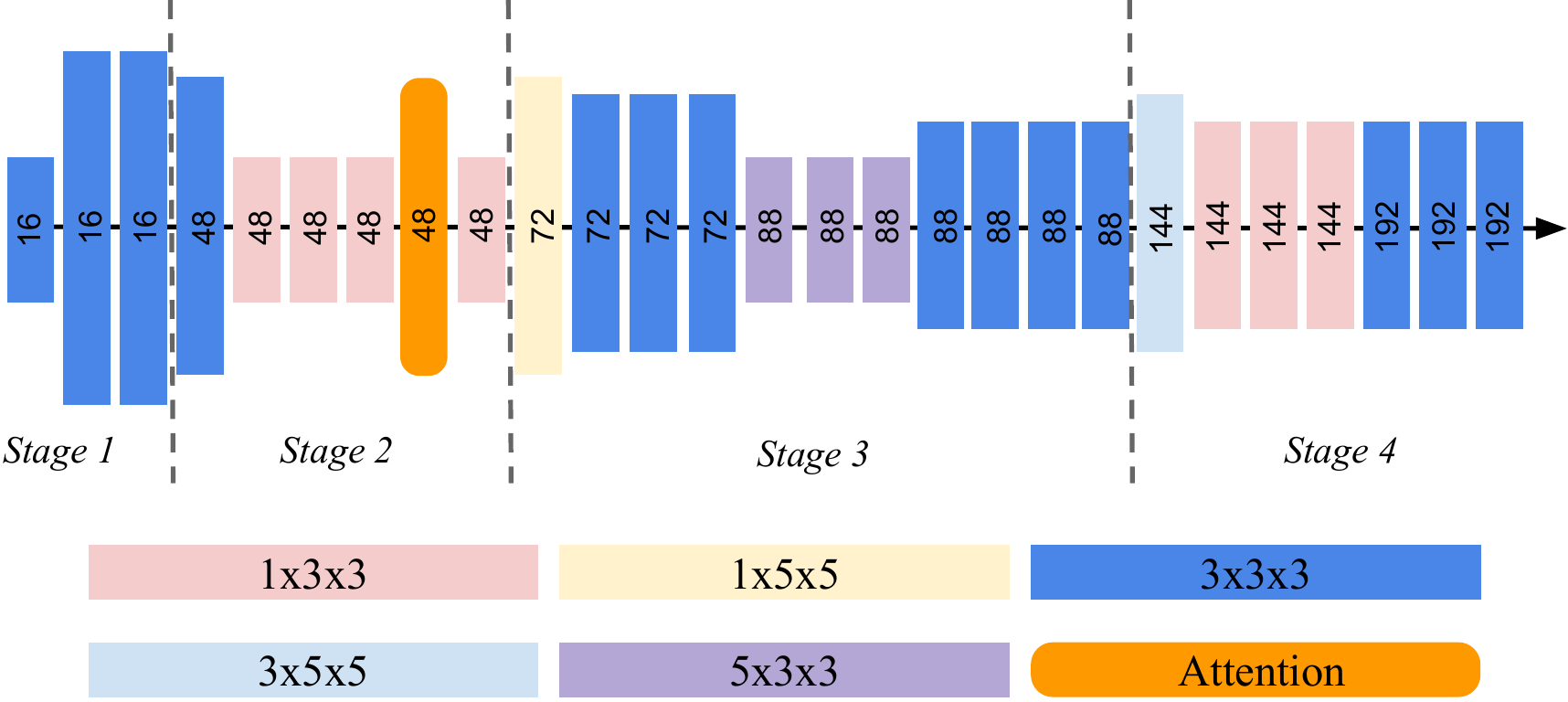}  
\caption{Visualization of the searched architecture. The number in each block represents the base channel of 3D MBConv block. The wider width of the block denotes the higher expansion ratio.}
    \label{fig:vis_arch}
\end{figure}
 Our searched \autoxthreed architectures are visualized in Figure~\ref{fig:vis_arch}. Regarding to the design of the backbone network, we find that $3 \times 3^2 $ convolution is widely adopted, while temporal kernel size 5 is hardly used. We also observe that $5 \times 5^2$ convolution is never used in our network, due to its high complexity.
 Interestingly, We also find that the expansion ratio is often higher at the first block of each stage (except stage 1).
 
 \vspace{-5pt}
\section{Conclusions and Future Work}
In this paper, we introduce a family of efficient AutoX3D models, by directly searching without using any architecture surrogate nor hand-crafted image architecture. The proposed efficient architectures are discovered through a finer-grained search space, where block type, filter number, expansion ratio and attention block are jointly searched. In addition, a fairness-aware search strategy is adopted to avoid collapse issue. Our best architecture achieves state-of-the-art performance both on Kinetics-400 dataset and Something-something-V2 dataset, \eg, $+$\textbf{1.3}\% accuracy compared to X3D-L, and $\times$\textbf{1.74} lower computational cost compared to X3D-XL, on Kinetics-400 validation set.


\clearpage

{\small
\bibliographystyle{ieee_fullname}
\bibliography{egbib}
}

\end{document}


\title{Supplementary for ``Auto-X3D: Ultra-Efficient Video Understanding via Finer-Grained Neural Architecture Search''}


\maketitle
\ifwacvfinal\thispagestyle{empty}\fi

\section{Searching Details}
We use SGD optimizer, with a cosine learning rate decay scheduler, to optimize model weights, and the architecture parameters are optimized with Adam optimizer. The learning rate of SGD optimizer is set to 1.6 and 0.025 for Adam optimizer. The searching takes 600 epochs, where the first 200 epochs are used for supernet warm up, where only the models weights are updated while keeping architecture parameters frozen. We use a batch size of 192.

\section{Training Details}
We initialize the derived model using~\cite{he2015delving}. Models are trained for 300 epoch. We adopt SGD as the optimizer with learning rate of 1.6, weight decay of $5\times10^{-5}$, momentum of 0.9, and a cosine scheduler of learning rate decay. The batch size is set to 192. We adopt dropout of 0.5 before the head of the network. Following~\cite{2019efficientnet}, we apply auto-augment~\cite{autoaugment} on each frame during training (consistent within each input video clip). For AutoX3D model family (S, M, L), we adopt scale jittering ranges of $[182, 228]$, $[256, 320]$, and $[356, 446]$, followed by a crop size of $160$, $224$, and $312$, respectively.

\section{General Fair Channel Selection Discussion}
\begin{figure}[!t]
\begin{center}
 \includegraphics[width=\linewidth]{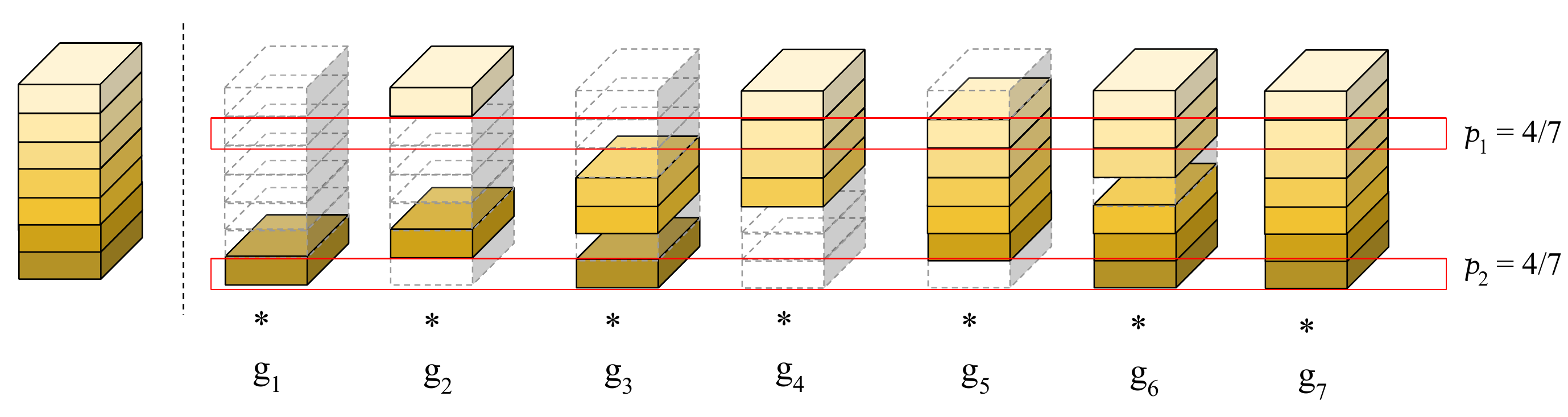}
\end{center}
  \caption{Another case of fair channel selection where $N = 7$ and $M=4$. More general cases are similar to it.}
\label{fig:random_fair}
\end{figure}

In the main text we give a specific example of fair channel selection that we choose 5 channel candidates. Here we discuss a general channel selection strategy. Considering $N$ channel candidates while $N$ has to be an odd number, the update probability of each filter is $p = \frac{M}{N}$ following fair selection strategy, where $M = \frac{N+1}{2}$. We give another example where we choose 7 channel candidates, shown in Figure \ref{fig:random_fair}. The update probability of each filter is $p=\frac{4}{7}$.


\section{More Results on Kinetics}
In this section, we present the results of AutoX3D models without attention blocks in Table \ref{tab:main_res}. We can observe that they still outperform X3D models with a large margin.
\begin{table}[t!]
\vspace{-3pt}
\centering
\resizebox{0.4\textwidth}{!}{
  \begin{tabular}{c|c|c|c}
     \multirow{2}{*}{Model} & Params & GFLOPS & Top-1  \\
      &(M) & $\times$views & Acc ($\%$) \\ 
    \specialrule{.15em}{.1em}{.1em}
      X3D-S~\cite{x3d} & 3.8 & 2.7 $\times$ 30& 73.3 \\
      AutoX3D-S w/o attention & 3.8 & 2.7 $\times$ 30 & 74.2 \\
      \textbf{AutoX3D-S (ours)} & 3.8 & 3.4 $\times$ 30& \textbf{74.7} \\ \hline
      
      X3D-M~\cite{x3d} & 3.8 & 6.2 $\times$ 30& 76.0 \\
      AutoX3D-M w/o attention & 3.8 & 6.2 $\times$ 30& 76.4 \\
      \textbf{AutoX3D-M (ours)} & 3.8 & 6.8 $\times$ 30& \textbf{76.7}\\\hline
      
      X3D-L~\cite{x3d} & 6.1 & 24.8 $\times$ 30& 77.5 \\
      AutoX3D-L w/o attention & 6.1 & 24.8 $\times$ 30& 78.3 \\
      \textbf{AutoX3D-L (ours)} & 6.2 & 27.8 $\times$ 30& \textbf{78.8}\\
     \bottomrule
  \end{tabular}
}
  \caption{\textbf{Comparisons with other NAS models on Kinetics-400.} AutoX3D and X3D models are evaluated using 30-views testing.}
\label{tab:main_res}
\end{table}

\section{Searched Architecture}
We present the architecture of our AutoX3D models in Table \ref{tab:arch}, where AutoX3D-L (Table \ref{tab:autox3d_l}) is obtained by naively stretching the depth of AutoX3D-S (Table \ref{tab:autox3d_s}) by 2$\times$.
\begin{table}[!t]
\begin{subtable}{\linewidth}
\centering
{\small
\resizebox{\linewidth}{!}{
\setlength{\tabcolsep}{1.2mm}{
    \begin{tabular}{lcccccc}
    \toprule
    Max Input & Block & \multirow{2}{*}{Expansion} & \multirow{2}{*}{Channel} & \multirow{2}{*}{Number} & Spatial & Att.   \\
    $C\times T \times S^2$ & Type &  &  &  &  Stride & Block \\
  \specialrule{.15em}{.1em}{.1em}      
    3$\times$$13$$\times160^{2}$ & data & 1 & 24 & 1 & 1 & -\\ \midrule
    3$\times$$13$$\times80^{2}$ & 1$\times3^{2}$ & 1 & 24 & 1 & 2 & -\\ \midrule
    28$\times$$13$$\times40^{2}$ &  t3\_s3  & 2.25 & 16 & 1 & 2  & - \\
    28$\times$$13$$\times40^{2}$ & t3\_s3 & 5.25 & 16 & 2 & 1 & - \\
    64$\times$$13$$\times20^{2}$ & t3\_s3 & 4.5 & 48 & 2 & 2 & - \\
    64$\times$$13$$\times20^{2}$ & t1\_s3 & 2.25 & 48 & 3 & 1 & GloRe \\
    132$\times$$13$$\times10^{2}$ & t1\_s5 & 4.5 & 72 & 2 & 2 & - \\
    132$\times$$13$$\times10^{2}$ & t3\_s3 & 3.75 & 72 & 3 & 1 & - \\
    132$\times$$13$$\times10^{2}$ & t5\_s3 & 2.25 & 88 & 3 & 1 & - \\
    132$\times$$13$$\times10^{2}$ & t3\_s3 & 3 & 88 & 3 & 1 & - \\
    264$\times$$13$$\times5^{2}$ & t3\_s5 & 3.75 & 144 & 2 & 2 & - \\
    264$\times$$13$$\times5^{2}$ & t1\_s3 & 3 & 144 & 2 & 1 & - \\
    264$\times$$13$$\times5^{2}$ & t3\_s3 & 3 & 192 & 3 & 1 & - \\
    \midrule
    432 & pool & - & 432 & 1 & - & - \\ \hline
    2048 & fc & - & 2048 & 1 & - & -\\
    \bottomrule
    \end{tabular}
    }
  }
}
\caption{\textbf{The architecture of \autoxthreed-S}.  \autoxthreed-M has the same architecture with \autoxthreed-S, except its input frame number is 16.}
\label{tab:autox3d_s}
\end{subtable}
\begin{subtable}{\linewidth}
\centering
{\small
\resizebox{\linewidth}{!}{
\setlength{\tabcolsep}{1.2mm}{
    \begin{tabular}{lcccccc}
    \toprule
    Max Input & Block & \multirow{2}{*}{Expansion} & \multirow{2}{*}{Channel} & \multirow{2}{*}{Number} & Spatial & Att.   \\
    $C\times T \times S^2$ & Type &  &  &  &  Stride & Block \\
  \specialrule{.15em}{.1em}{.1em}      
    3$\times$$16$$\times312^{2}$ & data & 1 & 24 & 1 & 1 & -\\ \midrule
    3$\times$$16$$\times156^{2}$ & 1$\times3^{2}$ & 1 & 24 & 1 & 2 & -\\ \midrule
    28$\times$$16$$\times78^{2}$ &  t3\_s3  & 2.25 & 16 & 1 & 2  & - \\
    28$\times$$16$$\times78^{2}$ & t3\_s3 & 5.25 & 16 & 5 & 1 & - \\
    64$\times$$16$$\times40^{2}$ & t3\_s3 & 4.5 & 48 & 4 & 2 & - \\
    64$\times$$16$$\times40^{2}$ & t1\_s3 & 2.25 & 48 & 6 & 1 & GloRe \\
    132$\times$$16$$\times20^{2}$ & t1\_s5 & 4.5 & 72 & 4 & 2 & - \\
    132$\times$$16$$\times20^{2}$ & t3\_s3 & 3.75 & 72 & 6 & 1 & - \\
    132$\times$$16$$\times20^{2}$ & t5\_s3 & 2.25 & 88 & 6 & 1 & - \\
    132$\times$$16$$\times20^{2}$ & t3\_s3 & 3 & 88 & 6 & 1 & - \\
    264$\times$$16$$\times10^{2}$ & t3\_s5 & 3.75 & 144 & 4 & 2 & - \\
    264$\times$$16$$\times10^{2}$ & t1\_s3 & 3 & 144 & 4 & 1 & - \\
    264$\times$$16$$\times10^{2}$ & t3\_s3 & 3 & 192 & 6 & 1 & - \\
    \midrule
    432 & pool & - & 432 & 1 & - & - \\ \hline
    2048 & fc & - & 2048 & 1 & - & -\\
    \bottomrule
    \end{tabular}
    }
  }
}
\caption{\textbf{The architecture of \autoxthreed-L}.  }
\label{tab:autox3d_l}
\end{subtable}
\caption{\textbf{The architecture of AutoX3D models.}}
\label{tab:arch}
\end{table}

\clearpage

{\small
\bibliographystyle{ieee_fullname}
\bibliography{egbib}
}